# Integration of Agile Ontology Mapping towards NLP Search in I-SOAS

Z. Ahmed[a,b], I. Tacheva[b,c]

[a] *University of Wuerzburg, Germany*
[b] *Vienna University of Technology, Austria*
[c] *Technical University Sofia, Bulgaria*

**Abstract**

In this research paper we address the importance of Product Data Management (PDM) with respect to its contributions in industry. Moreover we also present some currently available major challenges to PDM communities and targeting some of these challenges we present an approach i.e. I-SOAS, and briefly discuss how this approach can be helpful in solving the PDM community's faced problems. Furthermore, limiting the scope of this research to one challenge, we focus on the implementation of a semantic based search mechanism in PDM Systems. Going into the details, at first we describe the respective field i.e. Language Technology (LT), contributing towards natural language processing, to take advantage in implementing a search engine capable of understanding the semantic out of natural language based search queries. Then we discuss how can we practically take advantage of LT by implementing its concepts in the form of software application with the use of semantic web technology i.e. Ontology. Later, in the end of this research paper, we briefly present a prototype application developed with the use of concepts of LT towards semantic based search.

## 1. Introduction

Product data management (PDM) is the computer based system which electronically maintains the organizational technical and managerial data to take advantage in maintaining and improving the quality of products and followed development processes [11]. Major objectives of product data management are to improve the quality of products, improve team coordination, deliver products at the time, reduce engineering environment based problems, provide better and secure access to the configuration based information, prevent error creation and propagation by increasing customization of products, efficiently managing the large volumes of engineering data in reusable form generated by computer based systems and reusing design information.

In 1980s, in the begging of PDM proposition, the concept was new and not very welcomed by the industry of that time but with the passage of time now PDM is becoming more famous and widely in use of many multinational companies. Now a days, most of the PDM based applications are contributing in industry by providing engineering information management module control to access, store, integrate, secure, recover and manage information in data warehouse, distributed networked computer environment based infrastructure, resource management, information structure management, workflow control and system administration [1]. There are many PDM based application developers like Metaphase (SDRC) [7], SherpaWorks (Inso), Enovia (IBM), CMS (WTC), Windchill (PTC) [8], and Smarteam (Smart Solutions) [10].

Like many other communities of different fields the community of PDM System development is also struggling in solving one of currently available challenges i.e. Static and Unintelligent Search. In every PDM System a search mechanism is required and implemented to locate the user's needed information. But unfortunately still there is no such intelligent search mechanism is available which can process user's natural language based queries and can extract the most optimized results in minimum possible time in return. Keeping eyes on discussed problem in PDM System development, we can say, right now PDM community needs a new approach which can be very helpful in implementing the concepts of Natural Language Processing and capable of intelligently handling user's structured and unstructured natural language based requests, process and model the information for fast, optimized and efficient information retrieval mechanism.

**2. I-SOAS**

To take advantage in solving the problem of implementing of an intelligent human machine interface consisting of intelligent user system communication, meta data extraction out of unstructured data, semantic oriented information modeling, fast managed data extraction and final user end data representation an approach Intelligent Semantic Oriented Agent Based Search (I-SOAS) [1] has been proposed. The proposed conceptual architecture of I-SOAS is consists of four main sequential iterative components i.e. Intelligent User Interface (IUI), Information Processing (IP) [6], Data Management (DM) and Data Representation (DR), as shown in Figure 1.

IUI is responsible for the intelligent user system communication. IUI is proposed as a flexible graphical user interface capable of first analyzing the source of input, forwarding inputted data for further processing and responding back to the user with end results. Moreover IUI is supposed to be flexible enough so then it can be learned in shortest possible time and redesigned by user itself according to his need and wish. To implement the IUI as shown in Figure.1, IUI is divided into two main sub-categories i.e.Graphical User Interface and Communication Sources. Graphical User Interface is consists of the concept of three more sub-categories i.e. Intelligent, Flexible and Agent to intelligently handle the user's unstructured requests, provide multiple options to redesign the graphical user interface according to the ease of the user by user itself and perform internal architectural component's agent based communication. Where as in Communication Sources, first the corresponding user is supposed to be identified to enable the correct communication mode, if it is a digital system then electronic data communication mode will be enabled and if it is natural system then natural language based communication mode will be enabled [12].

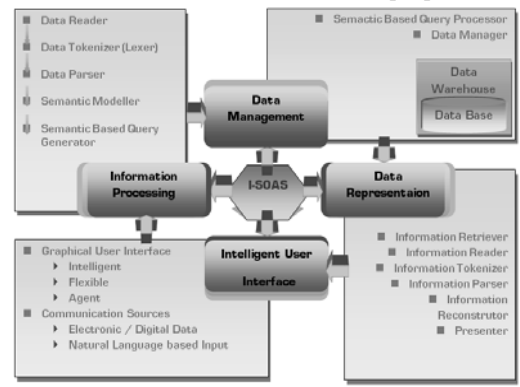

Figure. 1. I-SOAS [12]

IP unit is the most important component of the I-SOAS. The quality of performance of I-SOAS depends upon the accuracy in the results produced by this component. The overall job of IP is divided into five main iterative sequential steps i.e.Data reading, Tokenization, Parsing, Semantic Modeling, Semantic based query generation. The main concept behind the organization of these five steps is to first understand the semantic hidden in the context of natural language based set of instructions and generate a semantic information processable model for the system's own understanding and information processing [5]. In the first step, Data Reader is supposed to read and organized inputted data from IUI into initial prioritized instructions list. Then in the second step Data Tokenizer is supposed to tokenize instruction one by one, which are then treated in the third step by Data Parser for parsing and semantic evaluation with respect to the grammar of used natural language. Then in the fourth step Semantic Modeler is supposed to first filter the irrelevant semantic less data and then generate Meta data based semantic model. Then in the last and fifth step Semantic Based Query Generator is supposed to generate a new query used for further data storage and extraction of desired result.

DM is responsible for two main functions i.e. Semantic based Query Processing and Database Management. Semantic based query built in IP is treated by Semantic based Query Processor to generate SQL query to run in to database to store and extract the required information. The job of Data Manager is to manage the processes of SQL query building, data extraction and creation of new indexes and storage based on newly retrieved information [12]. DR is responsible for responding back to the user with finalized end results. This component consists of six sub components i.e. Information Retriever, Information Reader, Information Tokenizer, Information Parser, Information Reconstructor and Presenter. The job of this component is somehow similar to the job of IP, but major difference is of handling data and information. IP treats data to process but DR treats information. Required extracted and managed information from DM is passed to DR using Information Retriever, which simply read and organized by Information Reader without performing any analytical action except the prioritization of informative statements. Then using Information Tokenizer and Information Parser statements are tokenized and parsed, then using Information Reconstructor finalized formatted information is supposed to be built in user's used natural language based grammar. Finally Presenter presents the resultant information to IUI to respond back to the user [12].

The presented research in this paper, has been conducted for the implementation of one of the main modules i.e. Information Processing of a research project I-SOAS. The scope of this research is limited with respect to the construction of grammar rules. In section 3 of this research paper we give a description of the Natural Language Processing (NLP), in section 4 the ANTLR tool is briefly discussed, which later is used to write natural language grammar. In section 5 we present our newly constructed grammar and in section 6 we discuss mapping of Grammar into Ontology. In section 7 we briefly present the prototype of I-SOAS and conclude the discussion in section 8 of this research paper.

## 3. Natural Language Processing (NLP)

Today one of the most targeted problems in the field of artificial intelligence (computer science) is to make machine this much intelligent so then it can almost behave like a human being. Some of the behaviours of human beings have been accomplished during machine implementation e.g. now days machines can hear with the use of microphone, speak by producing sound, see with the use of cameras, smell with sensors but still there are some areas where this machine development is not completely successful and some of them are to understand natural language, learning from experience and making autonomous decisions in real time environment etc. In this research paper we will not discuss the moral problems which arise when it comes to machine development with intellectual capacities that rivals human beings but we will handle with the problems of how to make the machine understand the human languages. Natural Language Processing is one of the major step towards the development of field Artificial Intelligence as it deals with the propositions consisting of the production ability to implement an intelligent system which can not only process information but also can understand the user instructions in natural language e.g. English. Natural Language Processing performs two major tasks i.e. Natural language understanding and Natural language generation as shown in figure 2, to full fill the set goals of human machine communication implementation.

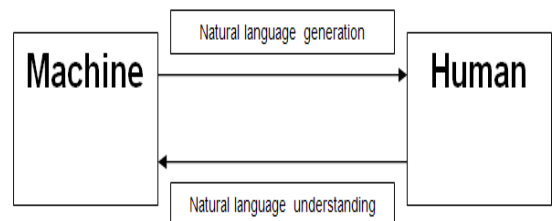

Figure.2. Human Maschine Communications

These tasks are implemented in two ways i.e. processes of converting the information from natural language to machine language and then vice versa converting the machine readable data to natural language based instructions. The main problem exists in between this communication is the structuring and restructuring of information with respect to the grammar of natural language and programming language involved.

Deep down at basic (low) level computers can only understand binary instructions to process with but here as we are dealing with the problem to make machine understand natural language, we need a natural language processor to translate languages from natural language to binary language and then back from binary to natural. To take advantage in the implementation of our targeted solution, we have used some programming language's concepts, whose compilers with the inclusion of assemblers are already doing the job of converting the information to binary level and then returning back the information after converting back to natural language statement (strings) from binary. To meet aforementioned goals, compiler consists of two main components i.e. Lexer and Parser. Lexer is basically a lexical analyser, based on the dictionary of the defined tokens of concerned language. Where as the Parser is the semantic analyzer based on the rules defined in the grammar.

NLP is one of the subfield of artificial intelligence, claiming the jobs of mainly analyzing, understanding and generating human (natural) languages, following the three step language conversion procedure as shown in figure 3. Likewise a programming language's compiler/interpreter, NLP consists of mainly three components i.e. Lexical Analyzer, Semantic Analyzer, and Translator. The lexical analyzer is also known as lexer, defining symbols or separate groups of symbols from the phrase [5]. It works like a filter program, which searches for certain characters by breaking actual character based statement(s) into tokens e.g. symbols, letters, digits, constants, reserved words, whitespace and comments etc., to perform certain defined tasks [4]. Furthermore it also regroups the input as series of characters with group significance i.e. tokens. Token is a symbol or group of symbols, don't have much sense at this level but they acquire meaning in the next step namely parser. The meaning of the tokens is given from the lexer rules. The stream of tokens generated by the lexer is received by the Semantic Analyzer to further process with.

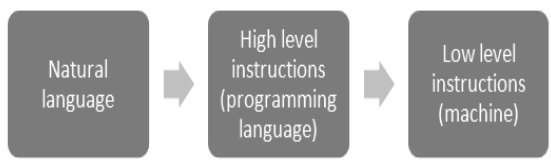

Figure.3. Language Conversion

The Semantic Analyzer is known as a Parser which receives the input source program and breaks these instructions into parts. It groups the tokens received from the lexer according to the given language grammar. If the grammar doesn't contains this group of tokens, the parser denies the successful processing otherwise proceed with.

There are two types of parser i.e. LL and LR parsers. The difference between them is in the derivations. The LL parsers construct a leftmost derivation of the input and LR constructs the rightmost derivation. This means that the LL parser replaces the left-most non terminal first and LR replaces the right most non terminal first, but both of them parses from left to right. Furthermore parser creates the sequences of tokens to put them into an Abstract Syntax Tree (AST) and makes one or many tables with information about the tokens or group of tokens i.e. symbol table, used to validate the types of the data.

## 4. ANLR

ANTLR is a tool, developed in 1983 [3] by Professor Terence Parr and his colleagues to write grammar for both lexer and parser. It is implemented in Java but it can generate source code in Java, C, C++, C#, Objective C, Python and Ruby. ANTLR use EBNF (Extended Backus-Naur Form) for the grammars, which is very formal way to describe the grammar. ANTLR provides a standard editor for grammar writing and generating lexer and parser. Till now this tool has been used for programming language's grammar writing but we are considering it for natural language processing by writing natural language's grammar and generating lexer and parser to make the machine understand it.

ANTLR has many belonging applications and opportunities to extensibilities. One of the biggest benefits is the grammar syntax; it is in EBNF form, which is a Meta syntax notation. Each EBNF rule has a left-hand side (LHS) which gives the name of the rule and a right-hand side (RHS) which gives the exact definition of the rule. Between the LHS and RHS there is the symbol ":" (colon), which separates the left from the right side and means "is defined as". One another benefit is the graphical grammar editor and debugger called ANTLRWorks, written by Jean Bovet and gives us the possibility to edit, visualize, interpret and debug any ANTLR grammar.

## 5. I-SOAS NLP Grammar

We have written and designed the grammatical view for the I-SOAS Lexer and I-SOAS Parser. This grammar is based on only English (Natural Language). At the moment, according to the scope of our research, the proposed grammar has been divided into three main categories i.e. A, B and C. A category is representing all English words belonging to the User (requesting for some results), B category is representing the English grammar structure and C is representing the main Object user is looking for. This grammar consists of 15 different lexer rules and 11 different parser rules in total. Please find generated grammar in section 9 Appendix of this research paper.

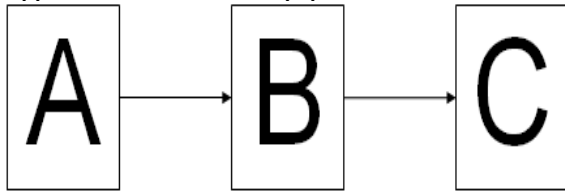

Figure.4. I-SOAS Lexer and Parser Grammar

**6. Mapping Grammar to Ontology**

One of the steps from Information Processing part in I-SOAS is the Semantic Modeler. I-SOAS Modeler makes a semantic model of the grammar and saves it into the Database (this is the connection between IP and DM). After the information is forwarded to the Resolver, which puts the relationships between the statements in the semantic model and also saves this into the Database. To make a conceptual model of the grammar we should use the ontology theory, which will describe its structure. Ontology clarifies the semantics of a conceptual modeling grammar [7]. In computer science the word ontology means to show one area of knowledge with conceptual scheme. The most used definition of ontology is from Gruber

> *"Ontology is a formal, explicit specification of a shared conceptualization" Gruber (1993)*

Formal means that the specification should be machine processable, explicit means that the elements should be simply defined and specification is an abstract model. According to Gruber the ontology is such a representation of a domain, where a set of objects and their relationships is described by a vocabulary [9]. There are different kinds of ontologies but we will use Natural Language Ontology (NLO) to provide the relationships between the different statements in our grammar [8]. There is also Domain Ontology (based on the knowledge of a particular domain) and Ontology Instance (which generates automatic object based web pages), all are connected to extract hidden semantic out of data.

This Ontology scheme is made from fields with the data and connections between the different objects, which declare the rules in the knowledge area. We can make an ontology scheme for our grammar to show how rules can be constructed by connecting the words. It includes a dictionary and logical relations between the terms and their meanings, the treatment of one term with others, different variations between them and a semantic model of the terms.

The ontology construction process begins with the parsed text from the natural language i.e. Words. The nouns in the ontology are called classes and the verbs are the properties, having different values for the different relationships. The first thing to do is making an oriented graph with all the classes and properties with the semantic relationships in between, then the ontology can be made with some of the ontology supporting languages like XML (eXtensible Mark-up Language), RDF (Resource Description Framework) and OWL (Web Ontology Language) [8] etc.

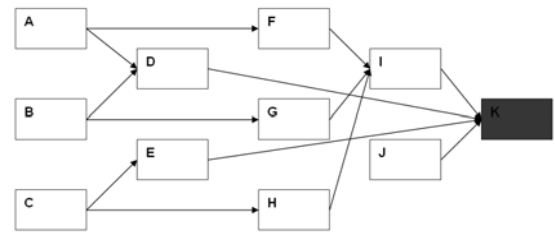

Figure.5. NLP Grammar Ontology Mapping

The map of our ontology should explain the relationships between the words in the grammar, what word comes after another. To answer the question "why", we will explain the English grammar. There are two types of verb tense that are used in our grammar i.e. Present Simple and Present Continuous. With them we have also two different persons and two different figures i.e singular and plural. This explains the statements from A to I i.e. here we are answering the questions: "Who", "What is he doing" or "What does he do" etc. Of course in the real search we can not use the personal pronoun because we can just say the action. That is why we have statements A D K, B D K, C E K, but also statements without A, B and C (which are the pronouns). D K, E, K is the example for Present Simple Tense, but we have it also for Present Continuous Tens. We have also statement J, which never uses a pronoun with itself (* J K).

In the picture of our ontology you can see the exact connections between the statements. There are in way just form left to right and there is no need to start always from A, we can start any sentence from where we want but follow the directions. The only mandatory statement is K, a key word, which defines what we want to search. In the following map scheme you can see the grammar with the connections between the words. As shown in figure 5, to produce ontology based search we have mapped the used grammatical view in Lexer and Parser in to an ontological view as shown in figure. There three main entities i.e. A, B and C. There are five main properties of A i.e. I, We, He, She and They, six properties of B i.e.am, are, is, looking/searching for, need(s)/want(s) and unknown and C has only one property K.

## 6. I-SOAS Prototype

I-SOAS Desktop Application as shown in figure 6. is capable of running as a stable application, taking natural language based instructions as input from user in the form of text file, and then lexing and parsing input text to build semantic based SQL queries to extract stored results from attached repositories.

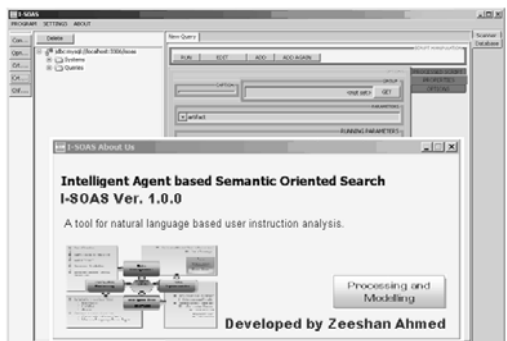

Figure.6. I-SOAS Desktop Prototype

## 6. Conclusion

In this research paper we have briefly addressed the importance of Product Data Management (PDM) along with currently available major challenge of natural language based search towards the Product Data Management System Development. Targeting the challenge, in this research paper, we have discussed a way for the implementation of a semantic based search by writing a natural language based grammar with the use of ANTLR and later mapped that grammar into ontology. As this is an ongoing research, we are hoping for a web based product data management system development in future, with the use of approach discussed in this research paper.